%% file: main_airoa.tex
\documentclass[10pt, a4paper]{article}
\usepackage{graphicx}
\usepackage{algorithm}
\usepackage{algpseudocode}
\usepackage{booktabs}
\usepackage{multirow}
\usepackage{amsmath} 
\usepackage{amssymb}  
\usepackage{makecell}
\usepackage{enumitem}
\usepackage{wrapfig}
\usepackage{array}
\usepackage{tabularx}
\usepackage{hybridreport}
\usepackage{setspace}

\usepackage[table,dvipsnames,svgnames]{xcolor}

\reporttype{}
\logo{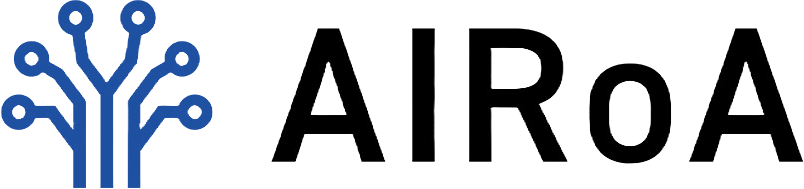}
\title{CounterAlign: Counterfactual Supervision for Vision-Language-Action Models}
\shorttitle{CounterAlign: Counterfactual Supervision for Vision-Language-Action Models}
\author{%
  Haru Kondoh\textsuperscript{1,2}
  \quad
  Kei Ota\textsuperscript{2}
  \quad
  Asako Kanezaki\textsuperscript{1,3,4}
  \quad
  Yueh-Hua Wu\textsuperscript{2}
}
\institution{%
  \textsuperscript{1}Institute of Science Tokyo
  \quad
  \textsuperscript{2}AI Robot Association (AIRoA)
  \\
  \textsuperscript{3}RIKEN AIP
  \quad
  \textsuperscript{4}Tohoku University
}
\date{June 2026}
\correspondence{\href{mailto:kondo.h.4aa3@m.isct.ac.jp}{\texttt{kondo.h.4aa3@m.isct.ac.jp}}}

\abstracttext{%
    Vision-Language-Action (VLA) models are typically trained with behavior cloning (BC) on expert demonstrations. However, BC provides only positive supervision for expert actions, without explicit negative supervision indicating which actions are instruction-inconsistent or otherwise inappropriate. Reinforcement learning (RL) can provide such corrective signals, but often relies on externally specified rewards or curated non-expert data, both of which are costly to obtain in robotics. We show that offline RL for VLA models need not rely on curated non-expert trajectories: successful expert demonstrations alone can be transformed into dense corrective supervision through instruction relabeling. Specifically, by pairing expert actions with mismatched alternative instructions, we synthesize counterfactual instruction–observation–action tuples from the dataset and combine them with adversarial discriminator training to learn an instruction-grounded reward model for offline RL, without collecting additional rollouts or annotations.
    On the robustness-focused LIBERO-PRO benchmark, our method improves robustness to object position and task perturbations over a strong state-of-the-art baseline. It also outperforms competitive baselines in real-robot experiments on the TX-G2 (compatible with AGIBot G2). More broadly, our results suggest that, for data-constrained VLA learning, extracting denser supervision from each demonstration can complement collecting additional data.
    Project page: \href{https://counteralign.airoa.io}{https://counteralign.airoa.io}.
}

\begin{document}
\maketitle


\begin{figure}[htbp]
  \vspace{-2.0em}
  \centering
  \includegraphics[width=\linewidth]{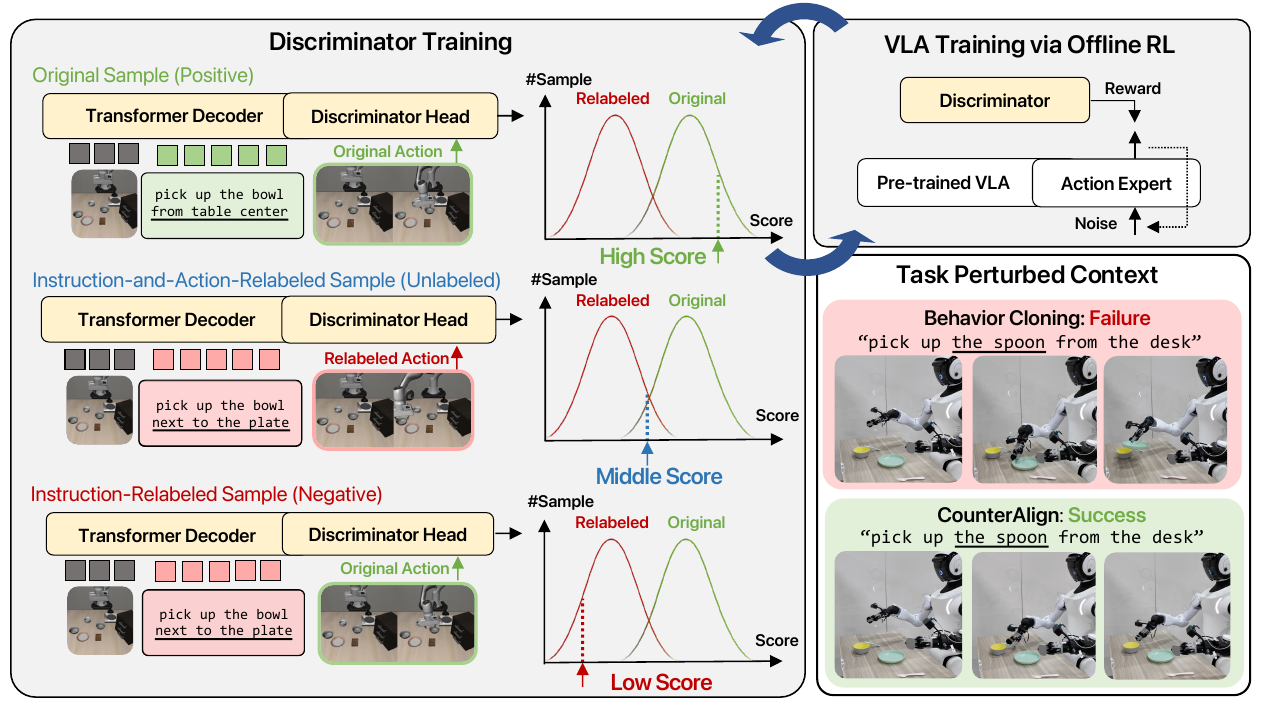}
  \vspace{-1.5em}
  \caption{
  Overview of \textbf{CounterAlign}. Through adversarial and relabeling-based discriminator training, CounterAlign converts expert demonstrations into an instruction-grounded reward. Relabeling creates counterfactual negative or unlabeled tuples that teach the reward model which instruction--action pairings are semantically inconsistent. The learned reward is then used to train a VLA policy with offline RL, improving robustness to object position and task perturbations without additional data collection or annotation.
  }
  \label{fig:teaser}
  \vspace{-2.0em}
\end{figure}

\input{text/introduction}
\input{text/related_work}
\input{text/method}

\input{text/experiments}

\input{text/conclusion}

Looking ahead, CounterAlign could be further strengthened by applying it to larger and more diverse robot datasets spanning a broader range of object categories, environments, embodiments, and linguistic variations. Another promising direction is to integrate counterfactual supervision with complementary data-collection or adaptation strategies, enabling future VLA systems to benefit from both denser supervision extracted from existing demonstrations and broader dataset coverage.

\separator

\subsection*{Acknowledgments}
{\small\color{TextMuted}%
    This paper is based on results obtained from a project, JPNP25015, commissioned by the New Energy and Industrial Technology Development Organization (NEDO).
}

\bibliographystyle{unsrt}
\bibliography{references}  

\input{text/supp}

\end{document}

%% file: text/introduction.tex
\section{Introduction}
\label{sec:introduction}

Recent progress in Vision-Language-Action (VLA) models has largely followed a simple recipe: collect more expert demonstrations and train larger policies with behavior cloning (BC) \cite{kim2024openvla,black2024pi0,black2025pi,ding2024quar,cheng2025navila,hirose2025omnivla}. Although effective, this recipe encounters two problems in robotics. First, robot data collection is expensive, embodiment-specific, and difficult to scale: unlike text or image corpora, every increase in coverage requires additional physical interaction under sensing, control, and safety constraints. Second, BC provides only positive supervision. It tells the policy what the expert did, but not which alternative behaviors would be wrong for the same instruction and observation. As a result, even strong VLAs remain brittle under paraphrases, object-position shifts, and scene variations \cite{zhou2025liberopro,fei25libero-plus}. If robot data are the bottleneck, then the key question is not only how to collect more demonstrations, but also how to extract more supervision from each one.

Reinforcement learning (RL), offline RL, and adversarial imitation learning provide a natural remedy because they can teach a model to distinguish desirable behavior from suboptimal behavior. In robotics, however, the supervision required for such learning is difficult to construct. Online RL requires informative rewards, diverse rollouts, and substantial simulator or hardware time, while remaining vulnerable to reward misspecification and reward hacking \cite{guo2025improving,zhai2025vision,intelligence2025pi,rlt2026}. Offline RL avoids additional interaction, but it still depends on a reward that faithfully captures instruction satisfaction and on a dataset with sufficiently informative behavioral support \cite{huang2025co,zhou2025reagentv,mao2026arm,zhang2026balancing}. Because offline RL cannot explore, policy improvement is constrained by the quality, diversity, and mixture of logged behavior; poor or weakly informative data can even degrade the learned policy. Thus, constructing the right mixture of expert and non-expert trajectories becomes a delicate dataset design problem. In practice, collecting and curating such suboptimal, failed, or corrective robot trajectories is itself expensive, and recent human-in-the-loop improvement pipelines still rely on substantial on-robot data collection together with expert interventions during deployment \cite{intelligence2025pi}.

To address this issue, we propose \emph{CounterAlign}, an offline RL framework that turns successful expert demonstrations into a source of corrective supervision through counterfactual relabeling.
The key observation behind CounterAlign is simple: a successful action chunk becomes informative about failure when paired with a semantically inappropriate instruction.
Concretely, by taking an expert action chunk and pairing it with a different instruction, we can synthesize mismatched instruction--observation--action tuples that reveal when a behavior is semantically inconsistent with the commanded task, without additional environment interaction, reward annotation, or corrective demonstrations.
We then extend this basic idea with action-side and jointly relabeled samples using similarity-based constraints, so that the reward model learns from nontrivial semantic near-misses rather than only easy mismatches.

However, relabeling is inherently ambiguous because different instructions can legitimately share similar or partially valid action chunks; treating all relabeled samples as negatives would inject systematic label noise and distort the reward. To address this, we train the relabeling discriminator using non-negative positive-unlabeled (nnPU) learning \cite{kiryo2017positive}, treating jointly relabeled samples as mixed unlabeled data rather than forcing them to be incorrect. 
In addition, we incorporate an adversarial learning objective that contrasts expert and policy-generated actions, encouraging the policy to remain close to expert-supported behaviors while improving semantic alignment.
Combined with an adversarial discriminator that contrasts expert and policy actions, this yields a dense reward that measures semantic alignment among language, observation, and action.

We then alternate between refining the discriminators and optimizing the VLA policy with offline RL using the reward induced by their outputs (see Fig.~\ref{fig:teaser}). The resulting framework requires no additional robot data collection or annotation, and increases the amount of useful supervisory signal extracted from each demonstration. Empirically, it improves robustness to object position and task perturbations and transfers to real robots, outperforming strong baselines on real systems.

The main contributions of this paper are summarized as follows.
\begin{itemize}[leftmargin=*]
    \item We propose a framework that converts expert demonstrations into informative counterfactual supervision for offline RL, without additional rollouts, annotations, or reward labels.
    \item We introduce an nnPU-based discriminator objective that explicitly handles the ambiguity of relabeled instruction-action pairs and produces a dense instruction-grounded reward.
    \item We integrate this reward into offline RL for VLA policy training and show improved robustness and generalization under object position and task perturbations while maintaining performance on the original training distribution and outperforming strong baselines in both simulation and real-robot experiments.
\end{itemize}

%% file: text/related_work.tex
\section{Related Work}
\label{related_works}

\noindent \textbf{Vision-Language-Action Models.}
Recent VLAs have made substantial progress in robotic manipulation by jointly learning from visual observations, language instructions, and actions \cite{kim2024openvla,black2024pi0,black2025pi,nvidia2025gr00tn1openfoundation}. However, recent benchmarks show that they remain brittle under paraphrased, compositionally novel, or semantically varied instructions \cite{zhou2025liberopro,fei25libero-plus,kim2026liberoparadiagnosticbenchmarkmetrics,wang2026libero}. We argue that this limitation stems not only from data scarcity, but also from how language is used in behavior cloning (BC): language conditions the policy, yet it does not serve as an explicit semantic constraint on the generated actions.
Recent work has therefore investigated reinforcement learning (RL) for VLA training. Simulation-based RL \cite{lu2025vlarlmasterfulgeneralrobotic,li2026simplevlarl,liu2025what,zang2025rlinf} suffers from sim-to-real gaps; world-model-based RL \cite{xiao2025worldenvleveragingworldmodel,sharma2026worldgymnasttrainingrobotsreinforcement,li2025vla,WMPO2025} can optimize over unrealistic states due to model errors; and real-world online RL \cite{guo2025improving,zhai2025vision,intelligence2025pi,rlt2026} incurs high data-collection costs and safety risks. Offline RL avoids additional interaction and enables policy improvement from fixed datasets \cite{huang2025co,zhou2025reagentv,mao2026arm,zhang2026balancing}, but existing approaches primarily rely on task-level or progress-oriented signals and do not directly address language grounding.

Several studies are more closely related to our goal of improving language-conditioned behavior. Glossop et al.~\cite{glossop2025cast} and Lee et al.~\cite{lee2026roboreward} generate counterfactual instructions, but this requires costly large-model relabeling and inherits the limitations of VLMs in spatial and physical reasoning. Kwok et al.~\cite{kwok2026scaling} align instructions and behaviors using contrastive learning, but in-batch negatives can become false negatives when instructions or behaviors are semantically similar. In contrast, our method filters highly similar negatives, treats jointly relabeled samples as unlabeled data within a positive-unlabeled learning framework, and integrates the learned reward into offline RL rather than using it only for test-time ranking.

\noindent \textbf{Reward Definition for VLAs.}
A central challenge in RL-based VLA training is how to define rewards that are informative and scalable. Many studies use binary success/failure rewards \cite{hu2025flare,zhang2025safevla,guo2025improving,tan2025interactive,zhao2025more,rlt2026,nakamoto2025steering,huang2025co,li2026simplevlarl} or sparse episode-level rewards with intermediate values \cite{intelligence2025pi,team2026gigabrain}. These rewards are simple, but they are inefficient and often require human annotation in real-world settings. Learned success classifiers \cite{xiao2026selfimproving,chen2025conrft} reduce the need for manual reward design, but they still capture task success rather than instruction--action semantic alignment, making them vulnerable to out-of-distribution instructions. Task-specific reward engineering is another option, but it does not scale to the broad task diversity targeted by VLAs \cite{mark2024policy,zhang2026balancing,wang2025vla,lyu2025reinforcement,shi2026beyond}.
To obtain denser rewards without additional annotation, prior work has used demonstration-derived progress signals \cite{shu2025rftf,zhang2025rewind,zhai2025vision,liu2026fly,bai2025evolve,mao2026arm,jiang2025irl,tan2025robo}. However, such signals assume that later states are more valuable and can be unreliable when demonstrations are suboptimal. Other methods rely on VLMs or video models to evaluate trajectories \cite{zhou2025reagentv,xiao2025world,sharma2026world,lee2026roboreward}, but they typically omit actions and inherit VLM weaknesses in spatial reasoning and semantic consistency \cite{zhou2026marvl}. Still others use discrepancies between world-model predictions and real trajectories \cite{li2025vla} or token-level differences from ground-truth actions \cite{kim2026roboalign}, but these signals are not trained to measure instruction--action alignment.
In this work, we instead learn a reward function that explicitly evaluates the semantic alignment between instructions and actions.

%% file: text/method.tex
\section{Method}
 \label{sec:method}

 \begin{figure}[t]
  \centering
  \includegraphics[width=\linewidth]{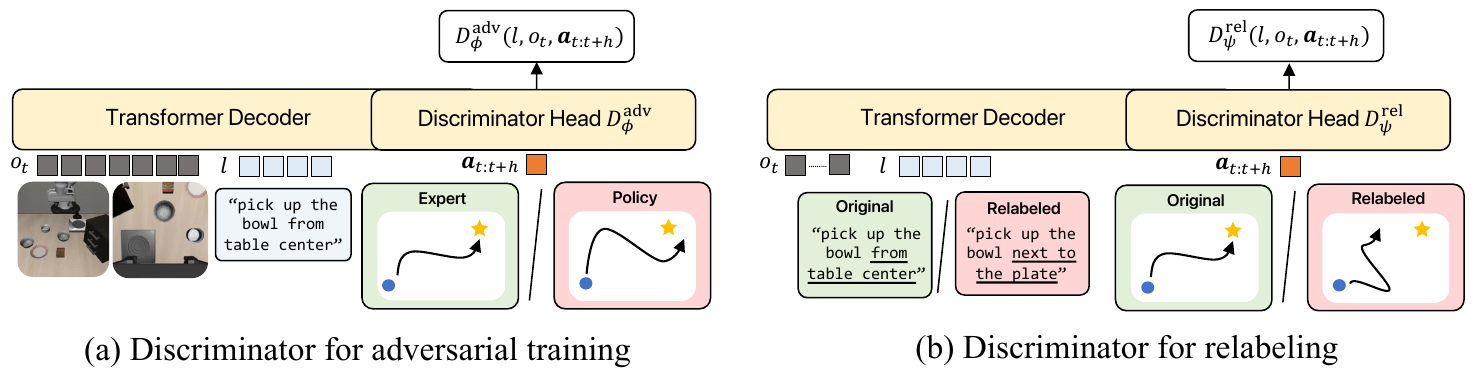}
  \caption{
    The two discriminators share the same architecture but are trained on different types of samples.
    (a) The adversarial discriminator is trained on original expert samples and samples whose actions are replaced with policy-generated actions.
    (b) The relabeling discriminator is trained on samples whose instructions or actions are replaced with other instructions or actions from the dataset.
  }
  \label{fig:proposed_method}
\end{figure}
 

\subsection{Problem Setting}

We consider sequential decision-making with observations $o$ (e.g., visual inputs, robot proprioceptions), language instructions $l$, and actions $a \in \mathbb{R}^d$. We are given an dataset $\mathcal{D} = \{(l^{i}, o^{i}_t, a^{i}_{t:t+h})\}$.
Our goal is to learn a VLA policy $\pi(a_{t:t+h}|o_t, l)$ that produces action trajectories semantically aligned with a given instruction $l$, without additional environment interaction.

\subsection{Learning Observation--Instruction--Action Alignment Reward}

We propose to train two discriminators (Fig.~\ref{fig:proposed_method}): an adversarial discriminator $D^{\mathrm{adv}}_\phi(l, o_t, a_{t:t+h})$ and a relabeling discriminator $D^{\mathrm{rel}}_\psi(l, o_t, a_{t:t+h})$, parameterized by $\phi$ and $\psi$, respectively. Both discriminators take as input an observation $o_t$, a language instruction $l$, and an action chunk $a_{t:t+h}$.

\noindent \textbf{Discriminator for Adversarial Training:}
The adversarial discriminator $D^{\mathrm{adv}}_\phi$ (Fig.~\ref{fig:proposed_method} (a)) determines whether a given action chunk comes from the expert or from the policy, as in GAIL~\cite{ho2016generative}. This discriminator can be interpreted as focusing on low-level action fidelity and evaluating whether the action faithfully follows the given instruction. In other words, it is sensitive even to slight deviations in the action.
A key difference from GAIL is that our method is trained entirely offline. We do not collect any additional data  during training. As positive samples for training the discriminator, we use dataset samples $\mathcal{D}_\mathrm{expert} = \{(l^{i}, o^{i}_t, a^{i}_{t:t+h})\}$. As negative samples, we use policy-generated actions $\mathcal{D}_\mathrm{policy} = \{(l^i, o_t^i, \pi(a_{t:t+h} \mid l^i, o_t^i))\}$. The discriminator $D^{\mathrm{adv}}_\phi$ is trained using the following loss:
\begin{equation}
\begin{aligned}
\mathcal{L}^\mathrm{adv}(\phi) =
&- \mathbb{E}_{
\mathcal{D}_\mathrm{expert}}
\left[ \log D^{\mathrm{adv}}_\phi(l, o_t, a_{t:t+h}) \right] 
- \mathbb{E}_{
\mathcal{D}_\mathrm{policy}}
\left[ \log \left(1 - D^{\mathrm{adv}}_\phi(l, o_t, a_{t:t+h}) \right) \right].
\end{aligned}
\label{eq:d_adversarial_loss}
\end{equation}
\noindent \textbf{Discriminator for Relabeling:}
The relabeling discriminator $D^{\mathrm{rel}}_\psi$ (Fig.~\ref{fig:proposed_method} (b)) evaluates, at a higher semantic level, whether a given action chunk follows the provided instruction. It is designed to be insensitive to small action deviations and instead performs a coarse-grained assessment of semantic consistency between the instruction and the action.
To train this discriminator, positive samples are drawn from the expert dataset $\mathcal{D}_{\mathrm{expert}} = \{(l^{i}, o^{i}_{t}, a^{i}_{t:t+h})\}$, as in the training of $D^{\mathrm{adv}}_\phi$. Here, $i$ indexes trajectories (episodes), trajectory $i$ consists of time steps $0, \ldots, T_i$, and $t$ denotes an arbitrary time index. For another trajectory $j$, we use $\tau$ to denote an arbitrary time index in trajectory $j$. Negative samples are constructed using two strategies: (1) instruction relabeling and (2) action relabeling.

\textbf{Instruction Relabeling.}
Instruction relabeling generates negative samples by replacing the original instruction $l^{i}$ in a training tuple $(l^{i}, o^{i}_{t}, a^{i}_{t:t+h})$ with a different instruction $l^{\mathrm{fake},i}$, resulting in $\mathcal{D}_{\mathrm{instr\_rel}} = \{(l^{\mathrm{fake},i}, o^{i}_{t}, a^{i}_{t:t+h})\}$.
Candidate instructions are selected from the dataset $\mathcal{D}$ such that $l^{\mathrm{fake},i} \neq l^{i}$.
Rather than selecting candidates uniformly at random, we select instructions whose similarity satisfies $\theta_{\mathrm{min}}^{l} < S_l((l^{i}, o^{i}_{t}, a^{i}_{t:t+h}), l^{\mathrm{fake},i}) < \theta_{\mathrm{max}}^{l}$.
This constraint avoids both trivial and ambiguous cases: overly similar instructions may be indistinguishable and lead to unstable training, whereas overly dissimilar instructions make the task too easy and fail to capture meaningful distinctions.
The instruction similarity $S_l$, together with its building blocks---the observation similarity $S_o$ and the action similarity $S_a$---is defined heuristically; we defer the precise formulations to the supplementary material.
The overall procedure for instruction relabeling is summarized in Algorithm~1 
in the supplementary material.

\textbf{Action Relabeling.}
Action relabeling generates negative samples by replacing the action chunk $a^{i}_{t:t+h}$ with another action chunk $a^{\mathrm{fake},i}$, resulting in $\mathcal{D}_{\mathrm{action\_rel}} = \{(l^{i}, o^{i}_{t}, a^{\mathrm{fake},i})\}$.
Candidate action chunks are selected from the dataset $\mathcal{D}$ such that $a^{\mathrm{fake},i} \neq a^{i}_{t:t+h}$.
Rather than selecting candidates uniformly at random, we select action chunks whose similarity satisfies $\theta_{\mathrm{min}}^{a} < S_a(a^{i}_{t:t+h}, a^{\mathrm{fake},i}) < \theta_{\mathrm{max}}^{a}$.
The overall procedure for action relabeling is summarized in Algorithm~2 
in the supplementary material.

\textbf{Joint Relabeling.}
We also construct jointly relabeled samples $\mathcal{D}_{\mathrm{unlabeled}} = \{(l^{\mathrm{fake},i}, o^{i}_t, a^{\mathrm{fake},i})\}$.
Candidate instructions are selected from the dataset $\mathcal{D}$ such that $l^{\mathrm{fake},i} \neq l^{i}$.
Rather than selecting candidates uniformly at random, we select instructions whose similarity satisfies $\theta_{\mathrm{min}}^{l} < S_l^o((l^{i}, o^{i}_{t}, a^{i}_{t:t+h}), l^{\mathrm{fake},i})$.
Here, $S_l^o$ denotes an observation-grounded variant of the instruction similarity that ignores the action term.
Candidate action chunks are selected from the samples associated with $l^{\mathrm{fake},i}$, i.e., from $\mathcal{S}(l^{\mathrm{fake},i})$, such that $a^{\mathrm{fake},i} \neq a^{i}_{t:t+h}$.
Rather than selecting candidates uniformly at random, we select action chunks whose proprioceptive similarity satisfies $\theta_{\mathrm{min}}^{p} < S_p(p^{i}_t, p^{\mathrm{fake},i})$, where $p^i_t$ denotes the proprioceptive observation corresponding to the robot state at time $t$ in the sample $(l^{i}, o^{i}_{t}, a^{i}_{t:t+h})$.
As with $S_l$, $S_o$, and $S_a$, both $S_l^o$ and $S_p$ are defined heuristically; their precise formulations are given in the supplementary material.
The overall procedure for joint relabeling is summarized in Algorithm~3 
in the supplementary material.

jointly relabeled samples are not guaranteed to be negative, since some instruction--action pairs may remain semantically consistent. We therefore treat them as unlabeled data and train $D^{\mathrm{rel}}_\psi$ using non-negative PU learning~\cite{kiryo2017positive}. The discriminator is trained with the following objective:
\begin{equation}
\begin{aligned}
\mathcal{L}^{\mathrm{rel}}(\psi)
&=
\lambda \mathcal{L}_{\mathrm{PN}}(\psi)
+
(1-\lambda) \mathcal{L}_{\mathrm{nnPU}}(\psi)
-
\alpha_{H} \mathcal{H}(D^{\mathrm{rel}}_\psi)
\end{aligned}
\label{eq:d_relabeling_loss}
\end{equation}
The last term is an entropy regularizer with coefficient $\alpha_{H} > 0$.
Here, $\mathcal{H}(D^{\mathrm{rel}}_\psi) \triangleq \mathbb{E}_{x \sim \mathcal{D}_{\mathrm{expert}} \cup \mathcal{D}_{\mathrm{neg}}}[H(D^{\mathrm{rel}}_\psi(x))]$ denotes the expected output entropy of the relabeling discriminator over the labeled data, where $H(p) = -p\log p - (1-p)\log(1-p)$ is the binary entropy function.
Without this regularizer, the discriminator tends to overfit to the training samples and collapse to saturated outputs, yielding degenerate reward signals that impede stable and meaningful policy optimization.

\subsection{Policy Training via Offline RL with Relabeling}

\noindent \textbf{Reward Definition:}
The RL reward is defined as a weighted sum of the log discriminator scores, analogous to GAIL:
\begin{equation}
    r_t(l^i, o_t^i, a_{t:t+h}^i)
    =
    (1-w)\log D^{\mathrm{adv}}_\phi(l^{i}, o^{i}_t, a^{i}_{t:t+h})
    +
    w\log D^{\mathrm{rel}}_\psi(l^{i}, o^{i}_t, a^{i}_{t:t+h}).
    \label{eq:reward_definition}
\end{equation}
Here, $w \in [0, 1]$ controls the relative contribution of the two discriminators.

\noindent \textbf{Relabeling for Critic Training:}
We additionally construct a relabeled dataset for critic training.
By incorporating more diverse instruction--observation--action samples into the critic objective, we aim to improve the critic’s evaluation of language-conditioned behavior.
The construction follows the same procedure as instruction relabeling for discriminator training
(Algorithm~1 
), but uses a different similarity threshold to select \emph{near-positive} samples rather than hard negatives.
Specifically, we construct $\mathcal{D}^{C}_{\mathrm{instr}} = \bigl\{(l^{j},\, o^{i}_{t},\, a^{i}_{t:t+h})\bigr\}$, where the relabeled instruction $l^{j}$ is selected such that $\theta^{l}_{\max} \leq S_l\!\left((l^{i}, o^{i}_{t}, a^{i}_{t:t+h}),\, l^{j}\right)$.
We restrict critic-side relabeling to instructions, omitting action relabeling.
Because our method is trained entirely offline, replacing the action chunk $a^{i}_{t:t+h}$ with one drawn from a different trajectory would leave the transition without a corresponding next observation $o^{i}_{t+h+1}$, which is required to form the bootstrapped target in the critic loss (Eq.~\ref{eq:q_value_loss}).
We then define the critic training dataset as $\mathcal{D}_{\mathrm{critic}} = \mathcal{D}_{\mathrm{expert}} \cup \mathcal{D}^{C}_{\mathrm{instr}}$.
We also explore actor-side relabeling as an ablation.
Further details are provided in the supplementary material.

\noindent \textbf{Training Algorithm:}
Our offline RL algorithm is based on Implicit Q-Learning (IQL)~\cite{kostrikov2022offline}.
We parameterize the Q-function as $Q_\theta$, the value function as $V_\xi$, and the actor as $\pi_\omega$.
To stabilize training, we additionally maintain a target network $Q_{\bar{\theta}}$.
In practice, we employ clipped double Q-networks~\cite{fujimoto2018addressing}, using two independent Q-networks $Q_{\theta_1}$ and $Q_{\theta_2}$ and replacing $Q_{\bar{\theta}}$ with $\min_{i\in\{1,2\}} Q_{\bar{\theta}_i}$ in the value and policy losses to mitigate overestimation bias.
We update the critics as follows:
\begin{equation}
\mathcal{L}^{\mathrm{Q}}(\theta) =
\mathbb{E}_{(l,o_t,a_{t:t+h},o_{t+h+1}) \sim \mathcal{D}_{\mathrm{critic}}}
\left[
\left(
r_t(l,o_t,a_{t:t+h})
+
\gamma V_\xi(l,o_{t+h+1})
-
Q_\theta(l,o_t,a_{t:t+h})
\right)^2
\right],
\label{eq:q_value_loss}
\end{equation}
\begin{equation}
\mathcal{L}^{\mathrm{V}}(\xi) =
\mathbb{E}_{(l,o_t,a_{t:t+h}) \sim \mathcal{D}_{\mathrm{critic}}}
\left[
L_2^\tau
\left(
Q_{\bar{\theta}}(l,o_t,a_{t:t+h}) - V_\xi(l,o_t)
\right)
\right],
\label{eq:value_loss}
\end{equation}
where $L_2^\tau(u) = |\tau - \mathbf{1}(u < 0)|\,u^2.$

For policy training, since the base policy $\pi_\omega$ is a flow-matching model, directly evaluating $\log \pi_\omega(a_{t:t+h} \mid l, o_t)$ is intractable for Advantage Weighted Regression (AWR)~\cite{peng2019advantage}.
Instead, we optimize the flow-matching loss weighted by the exponential of the advantage:
\begin{equation}
\mathcal{L}^{\mathrm{policy}}(\omega) =
\mathbb{E}_{\substack{(l,o_t,a_{t:t+h}) \sim \mathcal{D}_{\mathrm{expert}}\\s \sim \mathcal{U}[0,1],\; x_0 \sim \mathcal{N}(0,I)}}
\!\left[
\exp\!\left(\beta (Q_{\bar{\theta}}(l,o_t,a_{t:t+h}) - V_\xi(l,o_t)) \right)
\left\| v_\omega(x_s,s \mid l,o_t) - u_s \right\|^2
\right],
\label{eq:policy_loss}
\end{equation}
where $x_s = (1-s)\,x_0 + s\,a_{t:t+h}$ is the linear interpolant and $u_s = a_{t:t+h} - x_0$.
This objective up-weights the flow-matching loss for high-advantage actions and down-weights it for low-advantage actions, thereby biasing the policy toward higher-return behaviors without requiring likelihood evaluation.
The theoretical justification is provided in the supplementary material.

In CounterAlign, we iteratively train the discriminators ($D^{\mathrm{adv}}_\phi$, $D^{\mathrm{rel}}_\psi$), critic ($Q_\theta$, $V_\xi$), and actor ($\pi_\omega$).
The complete algorithm is provided in Algorithm~4 
in the supplementary material.

%% file: text/experiments.tex
\section{Experiments}
\label{sec:experiments}

\subsection{Simulation Setup}
Our objective is to examine whether a VLA model can go beyond simply memorizing a direct mapping from observations and language instructions to actions. Specifically, we investigate whether the model can interpret observations, understand the semantics of instructions, and determine the appropriate action accordingly. If a policy acquires such a semantically grounded action-generation capability, it is expected to exhibit robustness to perturbations in observations, object configurations, language expressions, and task settings.
We first train policies on the LIBERO dataset \cite{liu2023libero} and evaluate them on LIBERO-PRO \cite{zhou2025liberopro}.
LIBERO-PRO introduces several types of perturbation to each task suite, allowing a more fine-grained evaluation of generalization. In this work, we consider four perturbation types: Obj, which changes the visual appearance of objects; Pos, which changes object positions; Sem, which paraphrases the language instruction while preserving its meaning; and Task, which modifies the task to be performed. 
As the base VLA for CounterAlign, we use $\pi_{0.5}$ \cite{black2025pi}. All models are trained using the default configuration. Further details are provided in the supplementary material. For evaluation, we run 100 trials and calculate the success rate.

\subsection{Simulation Experiments}

\begin{table}[t]
    \setlength{\tabcolsep}{2.5pt}
    \centering
    \caption{
        Success rate comparisons on the LIBERO-PRO \cite{zhou2025liberopro}.
        CounterAlign yields its largest improvements in the Pos and Task settings, where the policy cannot succeed by simply replaying actions observed in the training data. 
    }
    \label{tab:libero_pro_results}
    \resizebox{\linewidth}{!}{%
    \begin{tabular}{lcccccccccccccccc}
        \toprule
        \multirow{2}{*}{Method}
        & \multicolumn{4}{c}{LIBERO-Spatial}
        & \multicolumn{4}{c}{LIBERO-Object}
        & \multicolumn{4}{c}{LIBERO-Goal}
        & \multicolumn{4}{c}{LIBERO-10}
        \\
        \cmidrule(lr){2-5}
        \cmidrule(lr){6-9}
        \cmidrule(lr){10-13}
        \cmidrule(lr){14-17}
        & Obj & Pos & Sem & Task
        & Obj & Pos & Sem & Task
        & Obj & Pos & Sem & Task
        & Obj & Pos & Sem & Task
        \\

        \midrule

        OpenVLA-OFT \cite{kim2025fine}
        & 0.30 & 0.14 & 0.65 & 0.0
        & 0.67 & 0.10 & 0.89 & 0.0
        & 0.85 & 0.0  & 0.44 & 0.03
        & 0.06 & 0.0  & 0.39 & 0.0
        \\

        VLA-Adapter \cite{wang2025vlaadapter}
        & 0.98 & 0.0 & \textbf{0.98} & 0.49
        & 0.89 & 0.0 & \textbf{0.99} & 0.08
        & 0.61 & 0.0 & 0.75 & 0.12
        & 0.47 & 0.0 & 0.91 & 0.10
        \\

        xVLA \cite{zheng2025x}
        & 0.90 & 0.0  & 0.69 & 0.39
        & \textbf{0.92} & 0.05 & 0.98 & 0.0
        & 0.72 & 0.10 & 0.94 & 0.08
        & 0.61 & 0.10 & 0.71 & 0.20
        \\
        
        $\pi_{0.5}$ \cite{black2025pi}
        & \textbf{0.99} & 0.53 & 0.97 & 0.55
        & 0.89 & 0.19 & 0.95 & 0.10
        & \textbf{0.90} & 0.29 & \textbf{0.95} & 0.17
        & \textbf{0.66} & 0.06 & 0.91 & 0.17
        \\

        \rowcolor{TableHighlight}
        CounterAlign
        & 0.98 & \textbf{0.60} & 0.97 & \textbf{0.63}
        & 0.80 & \textbf{0.51} & \textbf{0.99} & \textbf{0.26}
        & 0.82 & \textbf{0.41} & 0.94 & \textbf{0.46}
        & 0.64 & \textbf{0.11} & \textbf{0.97} & \textbf{0.29}
        \\
        \bottomrule
    \end{tabular}%
    }
\end{table}

The results on LIBERO-PRO are shown in Table ~\ref{tab:libero_pro_results}. In the Sem and Obj settings, we observe no substantial performance gap between the BC baseline and CounterAlign. One possible reason is that these settings tend to overestimate the performance of BC. In particular, in Sem and Obj, a policy can often achieve a high success rate simply by reproducing the same action sequences in response to observations similar to those seen during training, without necessarily understanding the semantics of the language instruction or selecting actions according to the situation.
Therefore, the competitive performance of BC in these settings does not by itself demonstrate semantic understanding; rather, it suggests that these evaluation settings can be solved to a large extent by a simple observation-to-action mapping.
%
%
In contrast, the Pos and Task settings require the policy to generate actions that differ from those observed during training, in response to changes in object configurations or task conditions. Therefore, these settings more directly evaluate the ability to interpret observations, understand the meaning of language instructions, and select actions appropriate to the current situation. CounterAlign yields its largest improvements in Pos and Task, demonstrating that learning a reward based on semantic alignment between language and action is effective in improving generalization under such perturbations.

\subsection{Ablation}
We conducted an ablation study to assess the contribution of each component (Table \ref{tab:libero_pro_ablation}). Even the basic offline RL configuration already outperforms behavior cloning in the most diagnostic Pos and Task settings, where the policy must adapt its actions to changes in object configurations or task requirements rather than simply replaying training trajectories. This suggests that our offline RL framework itself provides an effective policy-improvement mechanism for VLA models.
The full ablation suggests that most components improve average performance, whereas actor-side relabeling does not consistently help. In particular, relabeling for the discriminator and critic, entropy regularization, and jointly relabeled unlabeled samples improve robustness, while directly applying relabeling to the actor does not yield additional gains. We provide a more detailed analysis in the supplementary material.

\begin{table}[t]
    \setlength{\tabcolsep}{2.5pt}
    \centering
    \caption{
        Ablation study on LIBERO-PRO \cite{zhou2025liberopro}. Base algorithms are IQL~\cite{kostrikov2022offline} and advantage-weighted flow matching.
        Ent. denotes whether the entropy regularization is used in discriminator training, and Unlabel denotes whether unlabeled samples are used for discriminator training. D, C, and A indicate whether relabeling is applied to the discriminator, critic, and actor, respectively. 
    }
    \label{tab:libero_pro_ablation}
    \resizebox{\linewidth}{!}{%
    \begin{tabular}{ccccccccccccccccccccc}
        \toprule
        \multirow{2}{*}{Ent.}
        & \multirow{2}{*}{Unlabel}
        & \multirow{2}{*}{D}
        & \multirow{2}{*}{C}
        & \multirow{2}{*}{A}
        & \multicolumn{4}{c}{LIBERO-Spatial}
        & \multicolumn{4}{c}{LIBERO-Object}
        & \multicolumn{4}{c}{LIBERO-Goal}
        & \multicolumn{4}{c}{LIBERO-10}
        \\
        \cmidrule(lr){6-9}
        \cmidrule(lr){10-13}
        \cmidrule(lr){14-17}
        \cmidrule(lr){18-21}
        & & & &
        & Obj & Pos & Sem & Task
        & Obj & Pos & Sem & Task
        & Obj & Pos & Sem & Task
        & Obj & Pos & Sem & Task
        \\

        \midrule

        - & - & $\times$ & $\times$ & $\times$
        & 0.96 & 0.56 & \textbf{0.97} & 0.60
        & 0.86 & 0.49 & \textbf{1.0} & 0.22
        & 0.85 & 0.36 & 0.93 & 0.29
        & 0.62 & 0.09 & 0.96 & 0.14
        \\

        $\times$ & $\times$ & $\checkmark$ & $\times$ & $\times$
        & \textbf{1.0} & 0.59 & 0.96 & 0.57
        & \textbf{0.92} & 0.54 & 0.99 & 0.22
        & 0.84 & 0.34 & \textbf{0.98} & 0.32
        & 0.66 & 0.08 & 0.92 & 0.16
        \\

        $\checkmark$ & $\times$ & $\checkmark$ & $\times$ & $\times$
        & 0.98 & 0.55 & 0.94 & 0.65
        & 0.82 & \textbf{0.63} & 0.99 & 0.18
        & 0.87 & 0.38 & 0.97 & 0.41
        & \textbf{0.69} & 0.08 & 0.92 & 0.19
        \\

        $\checkmark$ & $\checkmark$ & $\checkmark$ & $\times$ & $\times$
        & 0.94 & 0.54 & \textbf{0.97} & \textbf{0.66}
        & 0.87 & 0.47 & 0.98 & 0.29
        & 0.83 & 0.37 & 0.96 & \textbf{0.47}
        & 0.63 & \textbf{0.12} & 0.89 & 0.20
        \\

        \rowcolor{TableHighlight}
        $\checkmark$ & $\checkmark$ & $\checkmark$ & $\checkmark$ & $\times$
        & 0.98 & \textbf{0.60} & \textbf{0.97} & 0.63
        & 0.80 & 0.51 & 0.99 & 0.26
        & 0.82 & \textbf{0.41} & 0.94 & 0.46
        & 0.64 & 0.11 & \textbf{0.97} & \textbf{0.29}
        \\

        $\checkmark$ & $\checkmark$ & $\checkmark$ & $\checkmark$ & $\checkmark$
        & 0.99 & 0.55 & \textbf{0.97} & 0.62
        & 0.90 & 0.62 & 0.97 & \textbf{0.35}
        & 0.80 & 0.36 & 0.97 & 0.33
        & 0.63 & 0.09 & 0.95 & 0.15
        \\

        \midrule
        BC & & & &
        & 0.99 & 0.53 & \textbf{0.97} & 0.55
        & 0.89 & 0.19 & 0.95 & 0.10
        & \textbf{0.90} & 0.29 & 0.95 & 0.17
        & 0.66 & 0.06 & 0.91 & 0.17
        \\

        \bottomrule
    \end{tabular}%
    }
\end{table}

\subsection{Real World Setup}
\begin{figure}[t]
  \centering
  \includegraphics[width=\linewidth]{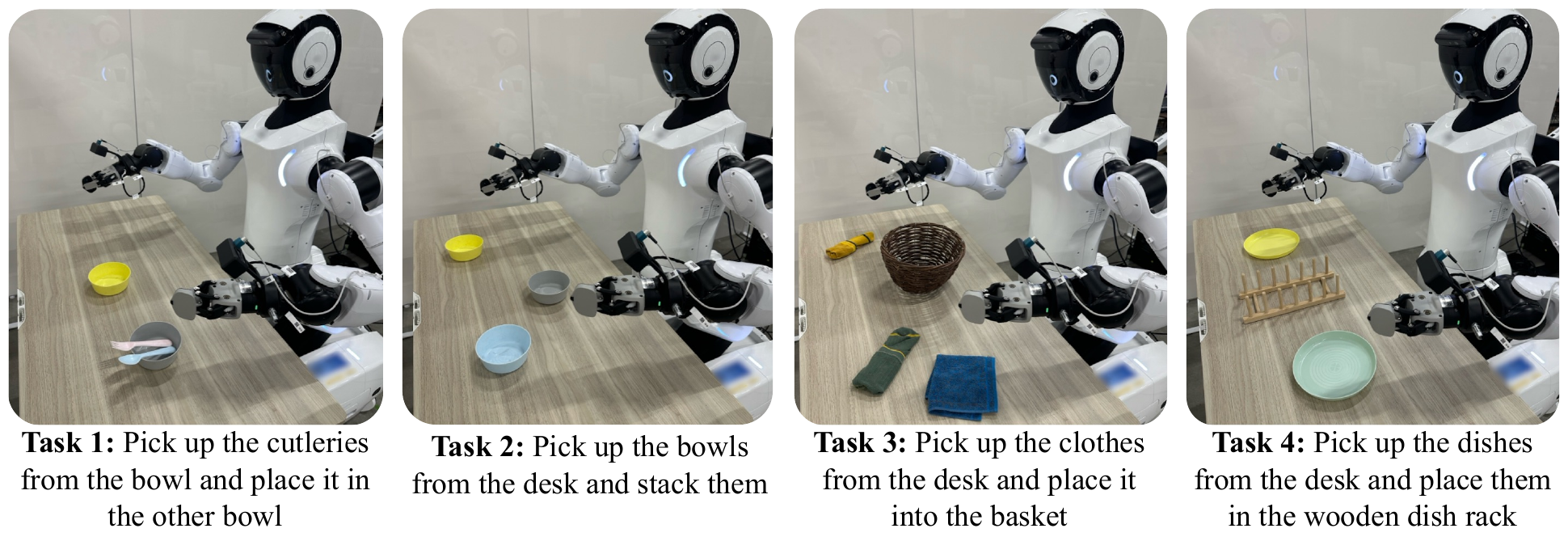}
  \caption{
  TX-G2 experimental setup. 
  }
  \label{fig:HSR_and_G2}
\end{figure}

We evaluated CounterAlign on a real robot, the TX-G2, as shown in Fig. \ref{fig:HSR_and_G2}.
We train models on a dataset consisting of four tasks. 
Each task is divided into 4 to 6 primitive tasks (e.g. ``pick up the light blue spoon from the grey bowl'').
For each task, we collect approximately 100 to 500 demonstrations. Further details are provided in the supplementary material.
To evaluate robustness in a real-robot setting, we conducted 10 trials. In each trial, perturbations were applied to the initial position and orientation of the target object. Among the 10 trials, 8 involved relatively small perturbations in the object pose, while the remaining 2 trials involved larger perturbations that were substantial enough to fall outside the training distribution.
We evaluate performance using the mean success rate across the primitive tasks in each task. 
Inference was run on an NVIDIA GeForce RTX 5070 GPU.

\subsection{Real World Experiments}

\begin{wraptable}{r}{0.40\linewidth}
    \vspace{-3.5em}
    \setlength{\tabcolsep}{2.5pt}
    \centering
    \caption{
        Comparisons on the TX-G2.
    }
    \vspace{-0.5em}
    \label{tab:g2_results}
    \resizebox{\linewidth}{!}{%
    \begin{tabular}{lcccc}
        \toprule
        Method
        & Task 1
        & Task 2
        & Task 3
        & Task 4 \\

        \midrule


        VLA-Adapter \cite{wang2025vlaadapter}
        & 0.0
        & 0.0
        & 0.0
        & 0.0 \\

        xVLA \cite{zheng2025x}
        & 0.0
        & 0.08
        & 0.05
        & 0.05 \\
        
        $\pi_{0.5}$ \cite{black2025pi}
        & 0.05
        & 0.48
        & 0.80
        & 0.60\\

        \rowcolor{TableHighlight}
        CounterAlign
        & \textbf{0.20}
        & \textbf{0.53}
        & \textbf{0.93}
        & \textbf{0.85} \\
        \bottomrule
    \end{tabular}%
    }
\end{wraptable}

Table \ref{tab:g2_results} shows that CounterAlign outperformed the baselines. In particular, even when perturbations were applied to the initial position and orientation of the object, the proposed method improved the performance. These results indicate that CounterAlign consistently achieves greater robustness than the baselines on the real robot.
\footnote{We also attempted to evaluate OpenVLA-OFT \cite{kim2025fine} in the real-robot setting. However, due to its large number of parameters, OpenVLA-OFT exceeded the available GPU memory.} 
%
%

\subsection{Qualitative Results}

\begin{figure}[t]
  \centering
  \includegraphics[width=\linewidth]{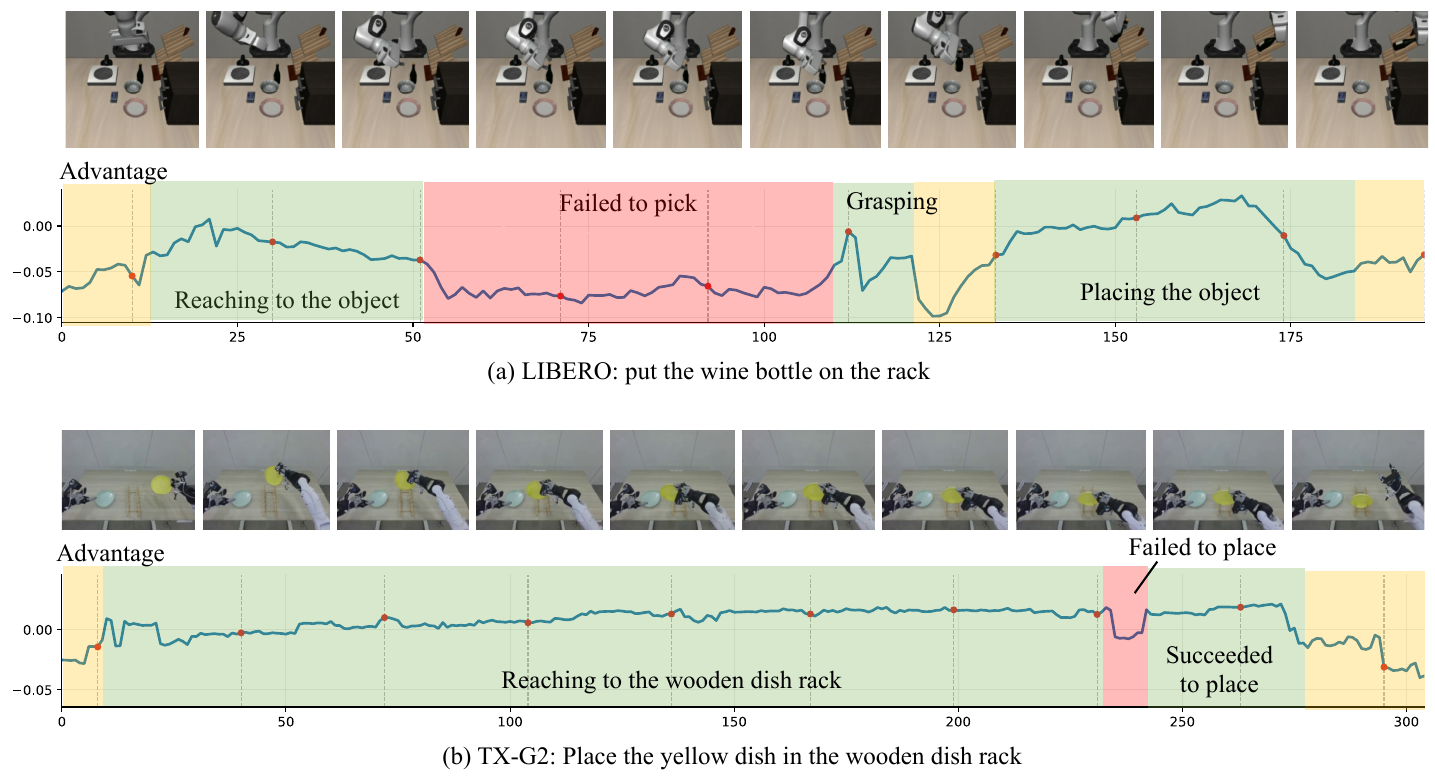}
  \caption{Trajectories from the LIBERO dataset and the TX-G2 dataset, along with the values of the learned advantage.}
  \label{fig:qualitative_result}
\end{figure}

Figure \ref{fig:qualitative_result} visualizes the learned advantage values along representative trajectories from LIBERO and the TX-G2 experiments. In both trajectories, the advantage is relatively low at the beginning and end of the episode. One possible explanation is that the task-relevant behavior is still ambiguous in the initial phase, while the final hand-lifting motion after object placement is only weakly related to the language instruction. We also observe that the advantage tends to decrease during failure segments, shown in red, and increase when the robot correctly follows the instruction, shown in green. In LIBERO, the advantage slightly decreases after the bottle is grasped, which may be because the dataset contains trajectories that place the bottle both on the rack and in the cabinet, making the intended target ambiguous at that moment. Overall, these results indicate that the learned advantage assigns higher values to trajectory segments strongly related to the instruction, and lower values to less relevant or unsuccessful segments.

%% file: text/conclusion.tex
\section{Conclusion}
\label{sec:conclusion}

We introduced CounterAlign, an offline reinforcement learning framework for improving the semantic robustness of Vision-Language-Action models. CounterAlign learns a reward function that explicitly evaluates the alignment among language instructions, observations, and action chunks, using adversarial and relabeling-based discriminators trained entirely from offline data. By integrating this learned reward into an IQL-style objective with advantage-weighted flow matching, the policy is encouraged to select actions that are semantically consistent with the current instruction and scene, rather than merely imitating spurious correlations in the dataset.
Experiments show that CounterAlign improves upon behavior cloning, particularly under position and task perturbations, where the policy must adapt its behavior to changes in the object position or task context. These gains carry over to real-robot experiments on TX-G2, where CounterAlign further outperforms competitive baselines.

%% file: text/supp.tex
\clearpage
\setcounter{page}{1}

\makeatletter
\title{Supplementary Material for\\CounterAlign: Counterfactual Supervision for Vision-Language-Action Models}
\author{}
\date{} 
\@maketitle
\makeatother

\setcounter{figure}{0}
\setcounter{table}{0}
\setcounter{section}{0}


This supplementary material provides additional details on the following topics:

\begin{enumerate}
  \item Definitions of similarity functions for relabeling
  \item Details of relabeling strategies
  \item Actor-side relabeling, an ablation variant
  \item The complete algorithm for the proposed method
  \item Theoretical justification of the advantage-weighted flow matching objective
  \item Model architectures and parameter counts for BC and the proposed method
  \item Training configuration
  \item Detailed ablation analysis
  \item Real-world datasets
  \item Additional experiments
\end{enumerate}

\section{Similarity Functions for Relabeling}
\label{sec:supp_similarity}

This section defines the similarity functions used to construct the relabeled datasets.
All similarity scores are heuristic and are computed offline from the expert dataset $\mathcal{D}_{\mathrm{expert}}$.

For an instruction $l$, we first define the associated set of observation–action pairs as
\[
\mathcal{S}(l) = \{(o^{k}_{t}, a^{k}_{t:t+h}) \mid l^{k} = l,\; (l^k, o^k_t, a^k_{t:t+h}) \in \mathcal{D}_\mathrm{expert} \}.
\]

\paragraph{Observation similarity $S_o$ and action similarity $S_a$.}
Let $(l^{i}, o^{i}_{t}, a^{i}_{t:t+h})$ be a reference sample, and consider another instruction $l^{j} \neq l^{i}$ with its associated set $\mathcal{S}(l^{j})$.
The observation similarity $S_o$ and action similarity $S_a$ are defined as
\begin{equation}
    S_o(o^{i}_{t}, o^{j}_{\tau}) =
    \frac{\left\langle \phi_o(o^{i}_{t}), \phi_o(o^{j}_{\tau}) \right\rangle}
    {\left\| \phi_o(o^{i}_{t}) \right\| \left\| \phi_o(o^{j}_{\tau}) \right\|},
\end{equation}
\begin{equation}
    S_a(a^{i}_{t:t+h}, a^{j}_{\tau:\tau+h}) =
    \prod_{k=0}^{h}
    \frac{\left\langle a^{i}_{t+k}, a^{j}_{\tau+k} \right\rangle}
    {\left\| a^{i}_{t+k} \right\| \left\| a^{j}_{\tau+k} \right\|},
    \label{eq:action_similarity}
\end{equation}
where $\phi_o(\cdot)$ denotes an observation feature extractor.

\paragraph{Instruction similarity $S_l$.}
Given the observation and action similarities above, we define instruction similarity as
\begin{equation}
    S_l((l^{i}, o^{i}_{t}, a^{i}_{t:t+h}), l^{j}) =
    \max_{(o^{j}_{\tau}, a^{j}_{\tau:\tau+h}) \in \mathcal{S}(l^{j})}
    S_o(o^{i}_{t}, o^{j}_{\tau}) \cdot S_a(a^{i}_{t:t+h}, a^{j}_{\tau:\tau+h}).
    \label{eq:instruction_similarity}
\end{equation}

\paragraph{Observation-grounded instruction similarity $S^o_l$.}
For joint relabeling, we also use a variant of instruction similarity that omits the action term:
\begin{equation}
    S_l^o((l^{i}, o^{i}_{t}, a^{i}_{t:t+h}), l^{j}) =
    \max_{(o^{j}_{\tau}, a^{j}_{\tau:\tau+h}) \in \mathcal{S}(l^{j})}
    S_o(o^{i}_{t}, o^{j}_{\tau}).
    \label{eq:instruction_obs_similarity}
\end{equation}

\paragraph{Proprioceptive similarity $S_p$.}
The proprioceptive similarity, which is used to filter candidate action chunks in joint relabeling, is defined as
\begin{equation}
S_p(p^{i}, p^{j}) =
\frac{\left\langle p^{i}, p^{j} \right\rangle}
{\left\| p^{i} \right\| \left\| p^{j} \right\|},
\label{eq:proprio_similarity}
\end{equation}
where $p$ denotes the proprioceptive component of the observation corresponding to the robot state.

\section{Details of Relabeling Strategies}

This section provides the full algorithmic details of the three relabeling strategies used to construct training data for the relabeling discriminator $D^{\mathrm{rel}}_\psi$.
All three strategies are applied offline to the expert dataset $\mathcal{D}_{\mathrm{expert}}$ and require no interaction with the environment.

\paragraph{Instruction Relabeling (Algorithm~\ref{alg:instruction_relabeling_for_d}).}
This strategy constructs hard-negative samples by replacing the original instruction $l^i$ with a semantically different instruction $l^j$ drawn from the dataset, while keeping the observation and action chunk fixed.
Candidate instructions are selected such that the instruction similarity score $S_l$ lies within a moderate range $(\theta^l_{\min}, \theta^l_{\max})$.
Instructions below $\theta^l_{\min}$ are too dissimilar to the original task context and would yield trivial negatives, whereas those above $\theta^l_{\max}$ are so similar that the relabeled pair may still represent a valid instruction--action alignment, leading to label noise.
The resulting dataset $\mathcal{D}_{\mathrm{instr\_rel}}$ therefore contains samples in which a plausible-sounding but mismatched instruction is paired with an action, encouraging the discriminator to learn fine-grained semantic alignment.

\begin{algorithm}[htbp]
\caption{Instruction Relabeling Strategy for Discriminator}
\label{alg:instruction_relabeling_for_d}
\begin{algorithmic}[1]
\Require Expert dataset $\mathcal{D}_{\text{expert}}$,
         similarity thresholds $\theta^l_{\min}$, $\theta^l_{\max}$
\Ensure Instruction-relabeled dataset $\mathcal{D}_{\text{instr\_rel}}$
\State $\mathcal{D}_{\text{instr\_rel}} \leftarrow \emptyset$
\For{each $(l^i, o^i_t, a^i_{t:t+h}) \in \mathcal{D}_{\text{expert}}$}
    \For{each $l^j \in \mathcal{D}_{\text{expert}}$ such that $l^j \neq l^i$}
        \State Compute $S_l\!\left((l^i, o^i_t, a^i_{t:t+h}),\, l^j\right)$
        \If{$\theta^l_{\min} < S_l\!\left((l^i, o^i_t, a^i_{t:t+h}),\, l^j\right) < \theta^l_{\max}$}
            \State $\mathcal{D}_{\text{instr\_rel}} \leftarrow \mathcal{D}_{\text{instr\_rel}} \cup \left\{(l^{j},\, o^i_t,\, a^i_{t:t+h})\right\}$
        \EndIf
    \EndFor    
\EndFor
\State \Return $\mathcal{D}_{\text{instr\_rel}}$
\end{algorithmic}
\end{algorithm}

\paragraph{Action Relabeling (Algorithm~\ref{alg:action_relabeling_for_d}).}
This strategy constructs hard-negative samples by replacing the action chunk $a^i_{t:t+h}$ with a different action chunk $a^j_{\tau:\tau+h}$ drawn from another expert sample, while keeping the original instruction and observation fixed.
Candidate action chunks are selected such that the action similarity score $S_a$ lies within $(\theta^a_{\min}, \theta^a_{\max})$.
This intermediate range ensures that the substituted action is physically distinguishable from the correct one, avoiding label ambiguity, yet not so dissimilar as to be trivially incorrect.
The resulting dataset $\mathcal{D}_{\mathrm{action\_rel}}$ therefore contains samples in which a plausible but incorrect action is presented for a given instruction and observation, training the discriminator to assess action-level appropriateness.

\begin{algorithm}[htbp]
\caption{Action Relabeling Strategy for Discriminator}
\label{alg:action_relabeling_for_d}
\begin{algorithmic}[1]
\Require Expert dataset $\mathcal{D}_{\text{expert}}$,
         similarity thresholds $\theta^a_{\min}$, $\theta^a_{\max}$
\Ensure Action-relabeled dataset $\mathcal{D}_{\text{action\_rel}}$
\State $\mathcal{D}_{\text{action\_rel}} \leftarrow \emptyset$
\For{each $(l^i, o^i_t, a^i_{t:t+h}) \in \mathcal{D}_{\text{expert}}$}
    \For{each $(l^j, o^j_\tau, a^j_{\tau:\tau+h}) \in \mathcal{D}_{\text{expert}}$ such that $a^j_{\tau:\tau+h} \neq a^i_{t:t+h}$}
        \State Compute $S_a\!\left(a^i_{t:t+h},\, a^j_{\tau:\tau+h}\right)$
        \If{$\theta^a_{\min} < S_a\!\left(a^i_{t:t+h},\, a^j_{\tau:\tau+h}\right) < \theta^a_{\max}$}
            \State $\mathcal{D}_{\text{action\_rel}} \leftarrow \mathcal{D}_{\text{action\_rel}} \cup \left\{(l^i,\, o^i_t,\, a^j_{\tau:\tau+h})\right\}$
        \EndIf
    \EndFor
\EndFor
\State \Return $\mathcal{D}_{\text{action\_rel}}$
\end{algorithmic}
\end{algorithm}

\paragraph{Joint Relabeling (Algorithm~\ref{alg:joint_relabeling_for_d}).}
This strategy simultaneously replaces both the instruction and the action chunk, producing samples $(l^{\mathrm{fake},i}, o^i_t, a^{\mathrm{fake}, i})$ whose labels are inherently ambiguous.
To make the relabeled combination physically plausible—and therefore genuinely ambiguous rather than trivially negative—we first filter candidate instructions using the observation-based instruction similarity score $S_l^o$, retaining only instructions whose associated demonstrations have similar robot states.
Among the action chunks associated with each candidate instruction, further filtering is then applied using the proprioceptive similarity score $S_p$, ensuring that the selected action originates from a robot configuration close to the current state $p^i_t$.
Because the resulting tuples may or may not be semantically consistent, they cannot be treated as definitive negatives; instead, they are used as \emph{unlabeled} data and incorporated into the non-negative PU learning objective.

\begin{algorithm}[htbp]
\caption{Joint Relabeling Strategy for Discriminator}
\label{alg:joint_relabeling_for_d}
\begin{algorithmic}[1]
\Require Expert dataset $\mathcal{D}_{\text{expert}}$,
         similarity thresholds $\theta^l_{\min}$, $\theta^p_{\min}$
\Ensure Jointly relabeled dataset $\mathcal{D}_{\text{unlabeled}}$
\State $\mathcal{D}_{\text{unlabeled}} \leftarrow \emptyset$
\For{each $(l^i, o^i_t, a^i_{t:t+h}) \in \mathcal{D}_{\text{expert}}$}
    \State $\mathcal{C}_l \leftarrow \emptyset$
    \For{each $l^j \in \mathcal{D}_{\text{expert}}$ such that $l^j \neq l^i$}
        \State Compute $S^o_l\!\left((l^i, o^i_t, a^i_{t:t+h}),\, l^j\right)$
        \If{$\theta^l_{\min} < S^o_l\!\left((l^i, o^i_t, a^i_{t:t+h}),\, l^j\right)$}
            \State $\mathcal{C}_l \leftarrow \mathcal{C}_l \cup \{l^j\}$
        \EndIf
    \EndFor
    \For{each $l^{\text{fake},i} \in \mathcal{C}_l$}
        \For{each $(o^j_\tau, a^j_{\tau:\tau+h}) \in \mathcal{S}(l^{\text{fake},i})$ such that $a^j_{\tau:\tau+h} \neq a^i_{t:t+h}$}
            \State Compute $S_p\!\left(p^i_t,\, p^j_\tau\right)$
            \If{$\theta^p_{\min} < S_p\!\left(p^i_t,\, p^j_\tau\right)$}
                \State $\mathcal{D}_{\text{unlabeled}} \leftarrow \mathcal{D}_{\text{unlabeled}} \cup \left\{(l^{\text{fake},i},\, o^i_t,\, a^j_{\tau:\tau+h})\right\}$
            \EndIf
        \EndFor
    \EndFor
\EndFor
\State \Return $\mathcal{D}_{\text{unlabeled}}$
\end{algorithmic}
\end{algorithm}

\section{Actor-Side Relabeling (Ablation Variant)}
\label{sec:supp_actor_relabeling}

In the main method, only the critic is trained with relabeled data,
\begin{equation}
    \mathcal{D}_{\mathrm{critic}}
    =
    \mathcal{D}_{\mathrm{expert}}
    \cup
    \mathcal{D}^{C}_{\mathrm{instr}},
\end{equation}
whereas the actor is trained only on the expert dataset.
For the ablation study, we also consider an extended variant in which the actor is trained with relabeled data.
This section describes how the relabeled datasets used in this variant are constructed.

Following the same relabeling procedures as those used for the discriminator
(Algorithms~\ref{alg:instruction_relabeling_for_d} and \ref{alg:action_relabeling_for_d}),
we use near-positive similarity thresholds rather than hard-negative thresholds.
In addition to $\mathcal{D}^{C}_{\mathrm{instr}}$, we construct an action-relabeled dataset
\begin{equation}
    \mathcal{D}^{\pi}_{\mathrm{act}}
    =
    \left\{
    \left(
        l^{i},
        o^{i}_{t},
        a^{j}_{\tau:\tau+h}
    \right)
    \right\},
\end{equation}
where the relabeled action chunk $a^{j}_{\tau:\tau+h}$ satisfies
\begin{equation}
    \theta^{a}_{\max}
    \leq
    S_{a}
    \left(
        a^{i}_{t:t+h},
        a^{j}_{\tau:\tau+h}
    \right).
\end{equation}
As with instruction relabeling, this threshold complements the one used for $\mathcal{D}^{D}_{\mathrm{act}}$ action chunks are selected to be highly similar to the original action chunk rather than moderately dissimilar.
This provides a near-positive supervision signal for the actor.

For this ablation variant, the actor training dataset is defined as
\begin{equation}
    \mathcal{D}_{\mathrm{policy}}
    =
    \mathcal{D}_{\mathrm{expert}}
    \cup
    \mathcal{D}^{C}_{\mathrm{instr}}
    \cup
    \mathcal{D}^{\pi}_{\mathrm{act}}.
\end{equation}
This dataset replaces $\mathcal{D}_{\mathrm{expert}}$ in the policy loss for this ablation variant.

\section{Complete Algorithm of the Proposed Method}
\label{sec:supp_full_algorithm}

Algorithm~\ref{alg:proposed_method} presents the complete training procedure for the proposed method.
The algorithm consists of two phases: offline dataset construction and iterative model training.
In Phase~1, the expert dataset $\mathcal{D}_{\mathrm{expert}}$ is augmented via instruction, action, and joint relabeling
(Algorithms~\ref{alg:instruction_relabeling_for_d}--\ref{alg:joint_relabeling_for_d})
to produce labeled negatives and unlabeled samples for discriminator training, as well as near-positive samples for critic training.
In Phase~2, the discriminators ($D^{\mathrm{adv}}_\phi$, $D^{\mathrm{rel}}_\psi$), critic ($Q_\theta$, $V_\xi$), and actor ($\pi_\omega$) are updated alternately for $K$ iterations.
The reward computed by the two discriminators guides the critic update, and the critic then provides the learning signal for the actor.

\begin{algorithm}[htbp]
\caption{The Proposed Method}
\label{alg:proposed_method}
\begin{algorithmic}[1]
\Require Expert dataset $\mathcal{D}_{\mathrm{expert}}$;
         thresholds $\theta^l_{\min}, \theta^l_{\max},\, \theta^a_{\min}, \theta^a_{\max},\, \theta^p_{\min}$;
        iterations $K$; hyperparameters $\beta, \gamma, \tau, \lambda$; EMA momentum $m$
\Ensure Trained policy $\pi_\omega$

\Statex
\Statex \textbf{// Phase 1: Dataset Construction}
\Statex

\Statex \textit{Instruction relabeling} \hfill (see Algorithm~\ref{alg:instruction_relabeling_for_d})
\State $\mathcal{D}^{D}_{\mathrm{instr}} \gets \{(l^j, o^i_t, a^i_{t:t+h})\}$
       \textbf{ s.t. } $\theta^l_{\min} < S_l((l^i, o^i_t, a^i_{t:t+h}), l^j) < \theta^l_{\max}$
       \Comment{hard negatives for $D^{\mathrm{rel}}_\psi$}
\State $\mathcal{D}^{C}_{\mathrm{instr}} \gets \{(l^j, o^i_t, a^i_{t:t+h})\}$
       \textbf{ s.t. } $\theta^l_{\max} \leq S_l((l^i, o^i_t, a^i_{t:t+h}), l^j)$
       \Comment{near-positive samples for critic training}

\Statex
\Statex \textit{Action relabeling} \hfill (see Algorithm~\ref{alg:action_relabeling_for_d})
\State $\mathcal{D}^{D}_{\mathrm{act}} \gets \{(l^i, o^i_t, a^j_{\tau:\tau+h})\}$
       \textbf{ s.t. } $\theta^a_{\min} < S_a(a^i_{t:t+h}, a^j_{\tau:\tau+h}) < \theta^a_{\max}$
       \Comment{hard negatives for $D^{\mathrm{rel}}_\psi$}

\Statex
\Statex \textit{Joint relabeling} \hfill (see Algorithm~\ref{alg:joint_relabeling_for_d})
\State $\mathcal{D}_{\mathrm{joint}} \gets \{(l^{\mathrm{fake},i}, o^i_t, a^j_{\tau:\tau+h})\}$
       \textbf{ s.t. } $\theta^l_{\min} < S^o_l((l^i, o^i_t, a^i_{t:t+h}), l^{\mathrm{fake},i})$
       and $\theta^p_{\min} < S_p(p^i_t, p^j_\tau)$
       \Comment{unlabeled samples for $D^{\mathrm{rel}}_\psi$}

\Statex
\Statex \textit{Aggregation}
\State $\mathcal{D}_{\mathrm{neg}} \gets \mathcal{D}^{D}_{\mathrm{instr}} \cup \mathcal{D}^{D}_{\mathrm{act}}$
       \Comment{labeled negatives for $D^{\mathrm{rel}}_\psi$}
\State $\mathcal{D}_{\mathrm{critic}} \gets \mathcal{D}_{\mathrm{expert}} \cup \mathcal{D}^{C}_{\mathrm{instr}}$

\Statex
\Statex \textbf{// Phase 2: Iterative Training}
\Statex

\State Initialize $D^{\mathrm{adv}}_\phi,\, D^{\mathrm{rel}}_\psi,\, Q_\theta,\, V_\xi,\, \pi_\omega$; set $\bar{\theta} \leftarrow \theta$
\For{$k = 1, \ldots, K$}

    \Statex \hspace{1em} \textit{Discriminator update}
    \State Sample $\mathcal{B}_{\mathrm{exp}} \sim \mathcal{D}_{\mathrm{expert}}$,\;
           $\mathcal{B}_{\mathrm{neg}} \sim \mathcal{D}_{\mathrm{neg}}$,\;
           $\mathcal{B}_{\mathrm{joint}} \sim \mathcal{D}_{\mathrm{joint}}$
    \State Sample $\mathcal{B}_{\pi} \gets \{(l^i, o^i_t, \hat{a}^i_{t:t+h})\}$,\;
           $\hat{a}^i_{t:t+h} \sim \pi_\omega(\cdot \mid l^i, o^i_t)$,\;
           $(l^i, o^i_t) \in \mathcal{B}_{\mathrm{exp}}$
    \State Update $D^{\mathrm{adv}}_\phi$ via $\mathcal{L}^{\mathrm{adv}}(\phi)$ using $(\mathcal{B}_{\mathrm{exp}},\, \mathcal{B}_{\pi})$
    \State Update $D^{\mathrm{rel}}_\psi$ via $\mathcal{L}^{\mathrm{rel}}(\psi)$ using $(\mathcal{B}_{\mathrm{exp}},\, \mathcal{B}_{\mathrm{neg}},\, \mathcal{B}_{\mathrm{joint}})$

    \Statex
    \Statex \hspace{1em} \textit{Critic update}
    \State Sample $\mathcal{B}_{\mathrm{critic}} \sim \mathcal{D}_{\mathrm{critic}}$
    \State Compute $r_t$ for each $(l, o_t, a_{t:t+h}) \in \mathcal{B}_{\mathrm{critic}}$
    \State Update $Q_\theta$ via $\mathcal{L}^{Q}(\theta)$ on $\mathcal{B}_{\mathrm{critic}}$
    \State Update $V_\xi$ via $\mathcal{L}^{V}(\xi)$ on $\mathcal{B}_{\mathrm{critic}}$
    \State Update target network: $\bar{\theta} \leftarrow (1-m)\bar{\theta} + m\theta$ \Comment{EMA with momentum $m$}

    \Statex
    \Statex \hspace{1em} \textit{Actor update}
    \State Sample $\mathcal{B}_{\mathrm{policy}} \sim \mathcal{D}_{\mathrm{expert}}$
    \State Update $\pi_\omega$ via $\mathcal{L}^{\mathrm{policy}}(\omega)$ on $\mathcal{B}_{\mathrm{policy}}$
    \Statex

\EndFor
\State \Return $\pi_\omega$
\end{algorithmic}
\end{algorithm}

\section{Theoretical Justification of the Advantage-Weighted Flow Matching Objective}
\label{sec:supp_fm_policy_loss}

\paragraph{AWR Optimal Policy.}
We first recall the key result of Peng et al.~\cite{peng2019advantage}.
Let $\mu(a_{t:t+h} \mid l, o_t)$ denote the behavior policy that generated the offline dataset $\mathcal{D}_{\mathrm{policy}}$, and let $d_\mu(l, o_t)$ denote the marginal distribution over context--observation pairs $(l, o_t)$ induced by $\mu$.
The KL-constrained policy improvement problem
\begin{equation}
    \max_{\pi}
    \;\mathbb{E}_{\substack{(l, o_t) \sim d_{\mu} \\ a_{t:t+h} \sim \pi(\cdot \mid l, o_t)}}
    \!\Bigl[ A(l, o_t, a_{t:t+h}) \Bigr]
    \quad\text{s.t.}\quad
    \mathbb{E}_{(l,o_t)\sim d_\mu}\!\left[D_{\mathrm{KL}}\!\bigl(\pi(\cdot \mid l, o_t) \,\|\, \mu(\cdot \mid l, o_t)\bigr)\right] \leq \epsilon
    \label{eq:kl_constrained}
\end{equation}
has the following closed-form optimal solution~\cite{peng2019advantage}:
\begin{equation}
    \pi^*(a_{t:t+h} \mid l, o_t)
    \;=\;
    \frac{1}{Z(l, o_t)}\,\mu(a_{t:t+h} \mid l, o_t)
    \exp\!\Bigl(\beta\, A(l, o_t, a_{t:t+h})\Bigr),
    \label{eq:awr_optimal}
\end{equation}
where $Z(l, o_t) = \int \mu(a \mid l, o_t)\exp\!\bigl(\beta A(l, o_t, a)\bigr)\,da$ is the partition function and $\beta > 0$ is the inverse temperature.

\paragraph{Derivation of the Advantage-Weighted Flow Matching Objective.}
We show that $\mathcal{L}^{\mathrm{policy}}(\omega)$ naturally arises by training the flow matching policy $\pi_\omega$ to imitate $\pi^*$ using samples from the offline dataset.
We show that $\mathcal{L}^{\mathrm{policy}}(\omega)$ arises naturally when the flow matching policy $\pi_\omega$ is trained to imitate $\pi^*$ using samples from the offline dataset.
If samples from $\pi^*$ were directly available, the corresponding flow matching loss would be
\begin{equation}
    \widetilde{\mathcal{L}}(\omega)
    =
    \mathbb{E}_{\substack{(l, o_t) \sim d_\mu,\; a_{t:t+h} \sim \pi^*(\cdot \mid l, o_t) \\[2pt] s \sim \mathcal{U}[0,1],\; x_0 \sim \mathcal{N}(0,I)}}
    \!\left[
    \bigl\| v_\omega(x_s, s \mid l, o_t) - u_s \bigr\|^2
    \right].
    \label{eq:ideal_fm_loss}
\end{equation}
Because $\pi^*$ is not directly accessible, whereas samples from $\mu$ are available in the offline dataset, we apply importance sampling to rewrite Eq.~\eqref{eq:ideal_fm_loss} as an expectation under $\mu$:
\begin{equation}
    \widetilde{\mathcal{L}}(\omega)
    =
    \mathbb{E}_{\substack{(l, o_t) \sim d_\mu,\; a_{t:t+h} \sim \mu(\cdot \mid l, o_t) \\[2pt] s \sim \mathcal{U}[0,1],\; x_0 \sim \mathcal{N}(0,I)}}
    \!\left[
    \frac{\pi^*(a_{t:t+h} \mid l, o_t)}{\mu(a_{t:t+h} \mid l, o_t)}
    \bigl\| v_\omega(x_s, s \mid l, o_t) - u_s \bigr\|^2
    \right].
    \label{eq:is_fm_loss}
\end{equation}
Substituting Eq.~\eqref{eq:awr_optimal} into the importance weight cancels the behavior policy terms, leaving only the partition function:
\begin{equation}
    \widetilde{\mathcal{L}}(\omega)
    =
    \mathbb{E}_{\substack{(l, o_t) \sim d_\mu,\; a_{t:t+h} \sim \mu(\cdot \mid l, o_t) \\[2pt] s \sim \mathcal{U}[0,1],\; x_0 \sim \mathcal{N}(0,I)}}
    \!\left[
    \frac{\exp\!\bigl(\beta\, A(l, o_t, a_{t:t+h})\bigr)}{Z(l, o_t)}
    \bigl\| v_\omega(x_s, s \mid l, o_t) - u_s \bigr\|^2
    \right].
    \label{eq:fm_with_partition}
\end{equation}
Since $Z(l,o_t)$ depends only on the context--observation pair and not on the action chunk $a_{t:t+h}$, it does not affect the relative weighting of action chunks for a fixed context.
Omitting this factor therefore preserves the advantage-induced preference among actions within each context, while changing only the marginal weighting of contexts in the supervised policy extraction objective.
Moreover, in offline continuous-control settings, explicitly estimating $Z(l,o_t)=\int \mu(a\mid l,o_t)\exp(\beta A(l,o_t,a))\,da$ is unreliable because the behavior policy is unknown and only a limited number of action samples is available for each context.
Following standard advantage-weighted policy extraction methods, which similarly use unnormalized exponential advantage weights~\cite{peng2019advantage,nair2020awac,wang2020critic}, we omit $Z(l,o_t)$ and obtain the practical surrogate
\begin{equation}
    \widetilde{\mathcal{L}}(\omega)
    =
    \mathbb{E}_{\substack{(l, o_t) \sim d_\mu,\; a_{t:t+h} \sim \mu(\cdot \mid l, o_t) \\[2pt] s \sim \mathcal{U}[0,1],\; x_0 \sim \mathcal{N}(0,I)}}
    \!\left[
    \exp\!\Bigl(\beta\, A(l, o_t, a_{t:t+h})\Bigr)
    \bigl\| v_\omega(x_s, s \mid l, o_t) - u_s \bigr\|^2
    \right].
    \label{eq:fm_drop_z}
\end{equation}
Finally, approximating the expectation under $d_\mu$ and $\mu(\cdot \mid l, o_t)$ by the empirical average over the offline dataset $\mathcal{D}_{\mathrm{policy}}$ gives
\begin{equation*}
\mathcal{L}^{\mathrm{policy}}(\omega)
\approx
\mathbb{E}_{\substack{(l,o_t,a_{t:t+h}) \sim \mathcal{D}_{\mathrm{policy}}\\s \sim \mathcal{U}[0,1],\; x_0 \sim \mathcal{N}(0,I)}}
\!\left[
\exp\!\left(\beta A(l,o_t,a_{t:t+h})\right)
\left\| v_\omega(x_s,s \mid l,o_t) - u_s \right\|^2
\right].
\end{equation*}

\paragraph{Discussion.}
This objective differs from Advantage Weighted Matching (AWM)~\cite{xue2025advantage}, which is designed for online RL and weights the flow matching loss by the raw advantage $A$ rather than by $\exp(\beta A)$.
In the online setting, negative advantage weights can be tolerated because the policy continually collects new samples and can recover from perturbations.
In the offline setting, however, assigning a negative weight to a data sample effectively reverses the gradient direction, pushing the policy \emph{away} from an in-distribution action and potentially toward out-of-distribution behavior not covered by the dataset.
Since $\exp(\beta A) > 0$ for all $A$, our weighting ensures that every update provides non-negative reinforcement of the target action, with the strength of reinforcement varying according to the advantage.
This avoids the instability caused by negative weights in the offline setting.

\section{Model Architecture and Parameter Counts of BC and the Proposed Method}
\label{sec:supp_parameter_count}

We compare the number of trainable parameters in the BC baseline and the proposed method.
The BC baseline contains approximately 3.4B parameters, whereas the proposed method contains approximately 12.7B parameters.
This increase stems from the additional discriminator and critic networks introduced by our method, while the policy itself remains unchanged from the BC baseline.

To minimize this overhead, we extensively share the VLM backbone across auxiliary networks, while keeping their lightweight expert heads separate so that each module can specialize for its own objective.
When modules use separate VLM backbones, their token embeddings are shared.
The detailed architectural choices are as follows:
\begin{itemize}
    \item \textbf{Discriminators.} The two discriminators $D^{\mathrm{adv}}_\phi$ and $D^{\mathrm{rel}}_\psi$ share a single VLM backbone, while each maintains its own discriminator expert head.
    \item \textbf{Critic.} The Q-network $Q_\theta$ and the V-network $V_\xi$ share a single VLM backbone, while each maintains its own critic expert head.
    \item \textbf{Double-Q network.} The two Q-networks used for the double-Q estimator share the same VLM backbone and differ only in their critic expert heads.
    \item \textbf{Target network.} In contrast, the target network maintains a fully separate VLM backbone from the online critic, as well as its own critic expert heads, so that the target estimates remain stable under EMA updates.
    \item \textbf{Policy.} The policy network $\pi_\omega$ retains exactly the same architecture as the BC baseline; no structural modifications are introduced.
\end{itemize}
Following the original $\pi_{0.5}$~\cite{black2025pi} setup, we fine-tune all parameters rather than using LoRA or other parameter-efficient adaptation methods.

\section{Training Configuration}

\noindent \textbf{LIBERO: }
We train our models using the default configuration for training $\pi_{0.5}$ on the LIBERO dataset.
We observed that training approximately converges within 10K--15K gradient steps, and therefore train all models for 15K gradient steps.
The batch size is set to 256.
Training is conducted on 8 NVIDIA H200 GPUs and takes approximately 28 hours.
The training time per gradient step is comparable to that of behavior cloning; however, because the proposed method introduces additional trainable components and substantially increases the total number of parameters, it requires correspondingly more GPU memory.
Specifically, while the behavior cloning baseline can be trained on a single NVIDIA H200 GPU, our method requires at least five H200 GPUs due to the additional memory overhead from the discriminator and critic networks.

\noindent \textbf{TX-G2: }
For all real-robot experiments, we use a training configuration similar to that used for LIBERO.
However, we train the models for 150K gradient steps and use the 150K-step checkpoint for evaluation.
The batch size is set to 64.
Training is conducted on 8 H200 GPUs and takes approximately 96 hours.

\section{Detailed Ablation Analysis}
We conduct an ablation study on LIBERO-PRO to evaluate the contribution of each component of the proposed method.

First, the results confirm the effectiveness of the underlying offline RL framework.
In this setting, the discriminator is trained through adversarial learning, the critic is optimized using IQL, and the actor is trained with advantage-weighted flow matching.
Even this basic configuration yields a noticeable improvement, demonstrating that offline RL-based policy improvement is effective compared with standard BC.

Next, the results show the importance of relabeling for the discriminator.
By training on samples in which either the language instruction or the action is replaced while the observation is kept fixed, the discriminator learns to evaluate whether the instruction and action are semantically consistent under the current observation.
The reward produced by this discriminator is then used to estimate the advantage, and the policy is trained to maximize this advantage.
As a result, the policy is encouraged not merely to imitate actions based on observations, but to select actions according to the meaning of the language instruction.
The entropy regularization term further improves performance, likely because it prevents the discriminator outputs from saturating near 0 or 1, thereby mitigating overfitting and providing a more stable and meaningful reward signal.
In addition, using unlabeled samples yields further improvements.
By incorporating ambiguous relabeled samples as unlabeled data rather than discarding them during discriminator training, the discriminator can better capture the semantic alignment between language and action.

Furthermore, applying relabeling not only to the discriminator but also to the critic leads to additional performance gains.
This suggests that the knowledge about language--action correspondence acquired through relabeling is more effectively transferred to the critic, which in turn provides more appropriate advantage signals to the actor.

In contrast, relabeling for the actor does not provide further improvement.
This may be because the learned advantage function is imperfect, and directly using relabeled actions as supervision for the actor can introduce noise into action-generation learning.
This result suggests that, in the proposed method, relabeling is more effective when incorporated indirectly into policy learning through the discriminator and critic, rather than being applied directly to the actor.

\section{Real-world Datasets}
\label{sec:real_world_datasets}

We evaluate our method on the TX-G2 dataset.
The dataset consists of several short-horizon tasks, each of which is further decomposed into a sequence of primitive actions.
For example, the short-horizon task ``Pick up the cutleries from the bowl and place them in the other bowl'' consists of primitive actions such as ``Pick up the light blue spoon from the grey bowl'' and ``Place the light blue spoon in the yellow bowl.''
This section reports detailed statistics for the short-horizon tasks and primitive actions in the real-world dataset.

The TX-G2 dataset consists of four short-horizon manipulation tasks: stacking bowls, placing clothes into a basket, transferring cutleries between bowls, and placing dishes into a wooden dish rack.
Tables~\ref{tab:g2_short_horizon} and~\ref{tab:g2_task_decomposition} summarize the short-horizon task demonstrations and primitive-action instances in the dataset.

\begin{table}[ht]
    \centering
    \caption{Short-horizon task statistics for the TX-G2 dataset.}
    \label{tab:g2_short_horizon}
    \begin{tabular}{lr}
        \toprule
        Short-horizon task & Num. demos. \\
        \midrule
        Pick up the cutleries from the bowl and place them in the other bowl. & 206 \\
        Pick up the bowls from the desk and stack them. & 129 \\
        Pick up the clothes from the desk and place them into the basket. & 514 \\
        Pick up the dishes from the desk and place them in the wooden dish rack. & 349 \\
        \midrule
        \textbf{Total} & \textbf{1,198} \\
        \bottomrule
    \end{tabular}
\end{table}

\begin{table}[ht]
    \centering
    \caption{Decomposition of short-horizon tasks into primitive actions for the TX-G2 dataset.}
    \label{tab:g2_task_decomposition}
    \begin{tabularx}{\linewidth}{p{0.38\linewidth}X}
        \toprule
        Short-horizon task & Primitive actions \\
        \midrule

        Pick up the cutleries from the bowl and place them in the other bowl.
        &
        \begin{enumerate}[leftmargin=*, nosep]
            \item Pick up the light blue spoon from the grey bowl.
            \item Place the light blue spoon in the yellow bowl.
            \item Pick up the pink fork from the grey bowl.
            \item Place the pink fork in the yellow bowl.
        \end{enumerate}
        \\

        \midrule

        Pick up the bowls from the desk and stack them.
        &
        \begin{enumerate}[leftmargin=*, nosep]
            \item Pick up the yellow bowl from the desk.
            \item Stack the yellow bowl on the grey bowl.
            \item Pick up the light blue bowl from the desk.
            \item Stack the light blue bowl on the yellow bowl.
        \end{enumerate}
        \\

        \midrule

        Pick up the clothes from the desk and place them into the basket.
        &
        \begin{enumerate}[leftmargin=*, nosep]
            \item Pick up the green socks from the desk.
            \item Place the green socks into the basket.
            \item Pick up the handkerchief from the desk.
            \item Place the handkerchief into the basket.
            \item Pick up the yellow socks from the desk.
            \item Place the yellow socks into the basket.
        \end{enumerate}
        \\

        \midrule

        Pick up the dishes from the desk and place them in the wooden dish rack.
        &
        \begin{enumerate}[leftmargin=*, nosep]
            \item Pick up the yellow dish from the desk.
            \item Place the yellow dish in the wooden dish rack.
            \item Pick up the green dish from the desk.
            \item Place the green dish in the wooden dish rack.
        \end{enumerate}
        \\

        \bottomrule
    \end{tabularx}
\end{table}

\section{Additional Experiments}

\noindent \textbf{CALVIN: }
In addition to LIBERO-PRO, we evaluate the proposed method on the commonly used CALVIN benchmark \cite{mees2022calvin}.
In this setting, the policy is trained on environments A, B, and C and evaluated on unseen environment D.
This benchmark tests whether a policy can execute known tasks involving known object categories and environmental elements in an unseen environment, where visual appearance, object arrangement, and language expressions may differ from those encountered during training.
Thus, CALVIN serves as a complementary evaluation to LIBERO-PRO, measuring visual, linguistic, and behavioral generalization from a different perspective.
We run 1,000 trials for CALVIN.

The results on CALVIN are shown in Table~\ref{tab:calvin_results}.
On the CALVIN ABC $\rightarrow$ D benchmark, the proposed method consistently outperforms the BC baseline, although the margin of improvement is relatively small.
Specifically, the average length increases from 3.93 for BC to 4.03 for our method, and slight improvements are observed at all evaluation stages in terms of the success rates for completing 1 to 5 consecutive tasks.
These results suggest that the BC baseline already achieves strong performance on CALVIN, leaving limited room for further improvement by the proposed method.

One possible reason for this result is that the CALVIN dataset is richer than LIBERO in terms of both scale and diversity.
While LIBERO contains only approximately 6.8K steps on average for each of its 40 tasks, CALVIN provides approximately 41.8K steps on average for each of its 34 tasks.
Moreover, the state-action distribution in CALVIN is also more diverse.
With sufficiently diverse training data, BC can achieve strong performance even in the unseen environment D.
As a result, the performance gap between BC and the proposed method, which explicitly learns semantic consistency among language, observations, and actions, becomes smaller.

Nevertheless, this result does not undermine the effectiveness of the proposed method.
Collecting large-scale datasets such as CALVIN, where episodes cover diverse object configurations and situations, is highly costly in real-world robot learning.
Therefore, the ability of the proposed method to achieve strong generalization even with limited and insufficiently diverse data can be regarded as one of its important advantages.
In this sense, the small improvement observed on CALVIN can be interpreted as a consequence of the sufficiently large and diverse dataset partially mitigating the limitations of BC.
The proposed method remains particularly useful in settings where collecting such large-scale and diverse data is expensive or impractical.

In the evaluation of completing 1 to 5 consecutive tasks, the success rate decreases monotonically as the number of required tasks increases.
This is because, even if the success rate for each individual task is approximately 90\%, completing multiple tasks in sequence requires the policy to avoid failure at every step, causing the overall success probability to decrease multiplicatively.
Thus, this decrease should be interpreted not as method-specific degradation but as a natural consequence of error accumulation in long-horizon evaluation.
Indeed, the proposed method outperforms BC at every level of consecutive task completion, indicating a small but consistent improvement in long-horizon task execution.

\begin{table}[t]
    \setlength{\tabcolsep}{2.5pt}
    \centering
    \caption{
        Comparison on the CALVIN ABC $\rightarrow$ D benchmark \cite{mees2022calvin}. The base VLA model is $\pi_{0.5}$ \cite{black2025pi}.
    }
    \label{tab:calvin_results}
    \begin{tabular}{lcccccc}
    \toprule
    \multirow{2}{*}{Algo.}
    & \multicolumn{5}{c}{Tasks completed in a row}
    & \multirow{2}{*}{Avg. len.} \\
    \cmidrule(lr){2-6}
    & 1 & 2 & 3 & 4 & 5 & \\

        \midrule

        BC
        & 0.906 & 0.835 & 0.776 & 0.733 & 0.682
        & 3.93 \\

        \rowcolor{TableHighlight}
        Ours
        & \textbf{0.913} & \textbf{0.848} & \textbf{0.804} & \textbf{0.757} & \textbf{0.703}
        & \textbf{4.03} \\

        \bottomrule
    \end{tabular}
\end{table}